\documentclass[journal]{IEEEtran}
\usepackage{graphicx}
\usepackage{xcolor}
\usepackage{flushend}

\usepackage{bm}

\usepackage{algorithm}
\usepackage{algorithmic}
\usepackage{amsmath}

\usepackage{hyperref}

\usepackage{booktabs}
\usepackage[small]{caption}
\usepackage{subcaption}

\begin{document}
\title{Biased Embeddings from Wild Data: Measuring, Understanding and Removing}

\author{Adam Sutton, Thomas Lansdall-Welfare, Nello Cristianini\\
Intelligent Systems Laboratory, University of Bristol, Bristol, BS8 1UB, UK\\
\{adam.sutton, thomas.lansdall-welfare, nello.cristianini\}@bris.ac.uk}

\maketitle              % typeset the header of the contribution
\begin{abstract}
Many modern Artificial Intelligence (AI) systems make use of data embeddings, particularly in the domain of Natural Language Processing (NLP). These embeddings are learnt from data that has been gathered ``from the wild'' and have been found to contain unwanted biases. In this paper we make three contributions towards measuring, understanding and removing this problem. We present a rigorous way to measure some of these biases, based on the use of word lists created for social psychology applications; we observe how gender bias in occupations reflects actual gender bias in the same occupations
in the real world; and finally we demonstrate how a simple projection can significantly reduce the effects of embedding bias. All this is part of an ongoing effort to understand how trust can be built into AI systems.
\end{abstract}
\section{Introduction\label{intro}}
With the latest wave of learning models taking advantage of advances in deep learning \cite{shrivastava2017learning,silver2016mastering,zhu2017unpaired}, Artificial Intelligence (AI) systems are gaining widespread publicity, coupled with a drive from industry to incorporate intelligence into all manner of processes that handle our private and personal data, giving them a central position in our modern-day society.

This development has lead to {\color{black} demand} for fairer AI, where we wish to establish trust in the automated intelligent systems by ensuring that systems represent us fairly and transparently. However, there has been growing concern about potential biases in learning systems \cite{angwin2016machine,flores2016false} which can be difficult to analyse or query for explanations of their predictions, leading to an increasing number of studies investigating the way black-box systems represent knowledge and make decisions \cite{fong2018net2vec,Jia17a,kahng2018cti,ribeiro2016should,samek2017evaluating}. Indeed, principled methods are now required that allow us to measure, understand and remove biases in our data in order for these systems to be truly accepted as a prominent part of our lives.

In the domain of text, many modern approaches often begin by embedding the input text data into an embedding space that is used as the first layer in a subsequent deep network \cite{cer2017semeval,nakov2016semeval}. These word embeddings have been shown to contain the same biases \cite{caliskan2017semantics}, due to the source data from which they are trained. In effect, biases from the source data, such as in the differences in representation for men and women, that have been found in many different large-scale studies \cite{flaounas2013research,Jia16,lansdall2017content}, carry through to the semantic relations in the word embeddings, which become baked into the learning systems that are built on top of them.% As such, \cite{bolukbasi2016man} proposed a method for potentially reducing the bias found in word embeddings by finding certain projections that relate to a bias one wishes to remove before orthogonally projecting the data to remove it from the embedding.

{\color{black}In this paper, we make three contributions towards addressing these concerns. First we propose a new version of the} Word Embedding Association Tests (WEATs) studied in \cite{caliskan2017semantics}, designed to demonstrate and quantify bias in word embeddings, {\color{black}which puts them} on a firm foundation by using the Linguistic Inquiry and Word Count (LIWC) lexica \cite{pennebaker2001linguistic} to systematically {\em detect} and {\em measure} embedding biases.

{\color{black}With this improved experimental setting, we} find that European-American names are viewed more positively than African-American names, male names are more associated with work while female names are more associated with family, and that the academic disciplines of science and maths are more associated with male terms than the arts, which are more associated with female terms.  {\color{black}Using this new methodology, we then find that there is a gender bias in the way different occupations are represented by the embedding. Furthermore, we use the latest official employment statistics in the UK, and} find that there is a correlation between the ratio of men and women working in different occupation roles and how those roles are associated with gender in the word embeddings. {\color{black} This suggests that biases in the embeddings reflect biases in the world.}

Finally, we look at {\color{black}methods of }{\em removing} gender bias from the word embeddings. {\color{black}Having established that there is a direction in the embedding space that correlates with gender, we use a simple orthogonal projection to remove that dimension from the embedding}. After projecting the embeddings, we investigate the effect on bias in the embeddings by considering the changes in associations between the words, demonstrating that the associations in the modified embeddings now correlate less to UK employment statistics among other things.

\section{Methodology}
\subsection{Word Embedding}
A word embedding is a mapping of words into an $n$-dimensional vector space. Given a corpus of text, a word embedding can be created that will translate that corpus into a set of semantic vectors representing each word. Each word that appears in the corpus will be represented by an $n$-dimensional vector to indicate its position within the embedding.

This embedding has a set of features that can be used in natural language processing methods. The nearest neighbours of a word will be other words that have similar linguistic or semantic meaning, when comparing words using a measurement such as cosine similarity. There are also linear substructures within the word embeddings that can explain how multiple words are related to each other, making it a useful preprocessing step for natural language processing applications.

A word vector for a given word will now be defined as $\bm{w}$. Word vectors are normalized to unit length for measurement:

\begin{equation}
\hat{\bm{w}} = \frac{\bm{w}}{||\bm{w}||}.
\end{equation}

All future analysis will be done using normalised word vectors, if vectors in the future are edited they will again be normalised to unit length.

\subsection{Comparison of embedded words\label{sec:wordComp}}
Two words vectors $\bm{w}_1$ and $\bm{w}_2$ within a vector space can be compared by taking the dot product of their words:

\begin{equation}
\langle \hat{\bm{w}_1} , \hat{\bm{w}_2} \rangle = \sum\limits_{i=1}^{n} \hat{\bm{w}}_{1,i} \cdot \hat{\bm{w}}_{2,i}.
\end{equation}

As both word vectors are normalized, this is equivalent to the cosine similarity between the two word vectors. A cosine similarity closer to $1$ means that the vectors are similar to each other, while a cosine similarity of $0$ means that the vectors are orthogonal to each other.

In addition to comparisons between individual word vectors, we can compare an individual word vector to a set of word vectors. This is done by finding the mean of the set, normalizing the resulting vector and calculating the dot product with the individual word vectors as follows:

\begin{equation}
\label{eqn:word2set}
\langle \hat{\bm{w}} , \hat{\bm{\mu}} \rangle = \sum\limits_{i=1}^{n} \hat{\bm{w}}_i \cdot \dfrac{\bm{\mu}_i}{||\bm{\mu}||}.
\end{equation}

The resulting calculation gives us how closely an individual word is associated with a larger set of words. This association can be used to assess how closely related a given word is to different topics or concepts within the embedding space.

\subsection{Removing Bias\label{sec:removingbias}}
To remove bias, first two vectors have to be identified that contain contrasting directions of the bias. These two vectors ($\bm{w}_1$ and $\bm{w}_2$) must be considered ``opposite'' of each other semantically, in terms of the bias that is required to be removed. The following method of debiasing is the same as presented in \cite{bolukbasi2016man}:

\begin{equation}
\bm{w}_b = \hat{\bm{w}}_1 - \hat{\bm{w}}_2,
\end{equation}
where the vector $\bm{w}_b$ will have the direction of bias in the embedding (for example, he and she are different genders and could potentially be used to capture a gender direction).

Using this bias direction, all word vectors can now have that component removed by projecting them into a space that is orthogonal to the bias vector:

\begin{equation}
\bm{w}_\bot = \hat{\bm{w}} - (\hat{\bm{w}} \cdot \hat{\bm{w}}_b^T) \cdot \hat{\bm{w}}_b,
\end{equation}
where $\bm{w}_\bot$ is the original word vector with the biased component removed. This resulting vector will now have the number of effective dimensions reduced to $n-1$, indicating that it is orthogonal to the bias vector. These orthogonal word vectors are required to be again be normalised for further analysis.

\section{Experiments}
In this paper, we conduct three experiments on semantic word embeddings. 
We first propose a new version of the Word Embedding Association Tests studied in \cite{caliskan2017semantics} by using the LIWC lexica to systematically detect and measure the biases within the embedding, keeping the tests comparable with the same  set of target words. We further extend this work using additional sets of target words, and compare sentiment across male and female names. Furthermore, we investigate gender bias in words that represent different occupations, comparing these associations with UK national employment statistics. In the last experiment, we use orthogonal projections \cite{bolukbasi2016man} to debias our word embeddings, and measure the reduction in the biases demonstrated in the previous two experiments.

\subsection{Data Description and Embedding\label{sec:dataDesc}}
In all of our experiments, the first step is to obtain semantic vectors from a word embedding that we wish to analyse. We use GloVe embeddings \cite{pennington2014glove}, pre-trained using a window size of $10$ words on a combination of Wikipedia from 2014, and the English Gigaword corpus \cite{parker2011english}, where each of the $400{,}000$ words in the vocabulary for this embedding are represented by a $300$-dimensional vector. These vectors capture, in a quantitative way, the nuanced semantics between words necessary to perform meaningful analysis of words, reflecting the semantics found in the underlying corpora used to build them.

The Wikipedia data includes the page content from all English Wikipedia pages as they appeared in 2014 when a snapshot was taken. The English Gigaword corpus is an archive of newswire text data from seven distinct international sources of English newswire covering several years up until the end of 2010 \cite{parker2011english}.

\begin{figure*}[t!]
\captionsetup{justification=centering}
    \centering
    \begin{subfigure}[t]{0.3\textwidth}
        \centering
        \includegraphics[width=\textwidth,height=\textwidth]{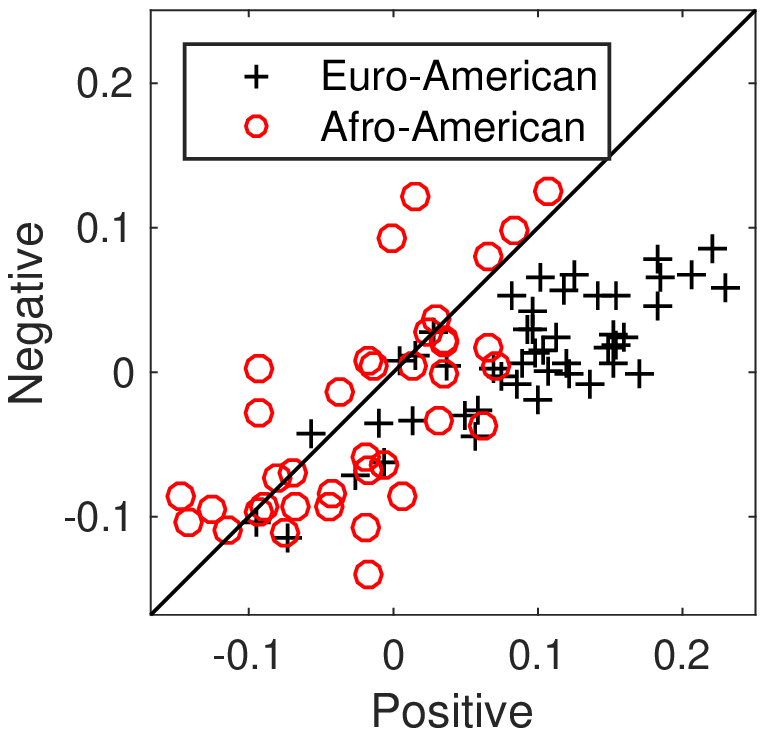}
        \caption{{\centering Association of European and African-American Names with Sentiment \label{fig:AfriAmer}}}
    \end{subfigure}%
    ~ 
    \begin{subfigure}[t]{0.3\textwidth}
        \centering
        \includegraphics[width=\textwidth,height=\textwidth]{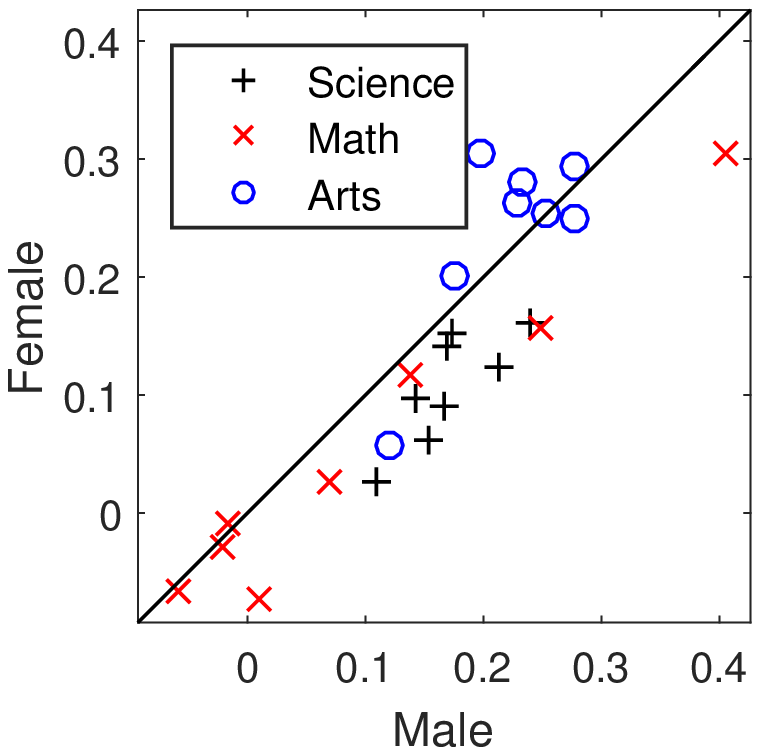}
        \caption{{\centering Association of Subject Disciplines with Gender \label{fig:ArtSciMath}}}
    \end{subfigure}
    ~ 
    \begin{subfigure}[t]{0.3\textwidth}
        \centering
        \includegraphics[width=\textwidth,height=\textwidth]{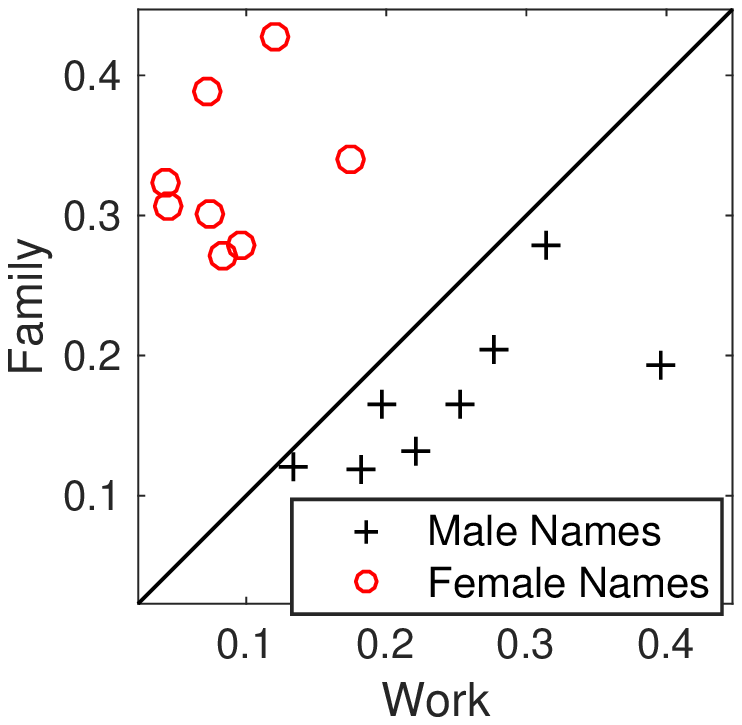}
        \caption{\label{fig:genderWorkFamily}{\centering Association of Gender with Career and Family}}
    \end{subfigure}
    \\\vspace{0.6cm}
    \begin{subfigure}[t]{0.3\textwidth}
        \centering
        \includegraphics[width=\textwidth,height=\textwidth]{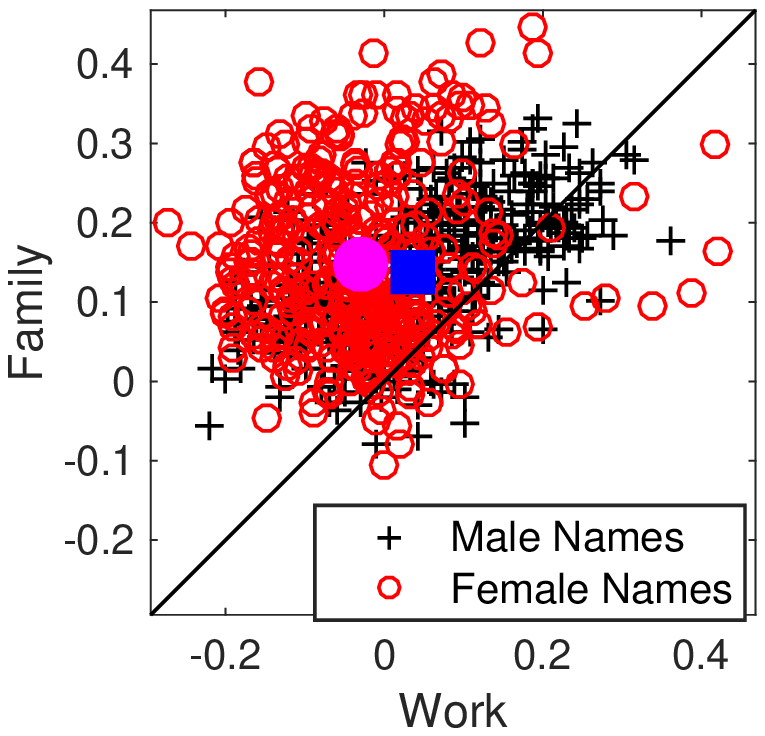}
        \caption{\label{fig:babyWorkFamily}{\centering Extended Association of Gender with Career and Family}}
    \end{subfigure}
    ~
    \begin{subfigure}[t]{0.3\textwidth}
        \centering
        \includegraphics[width=\textwidth,height=\textwidth]{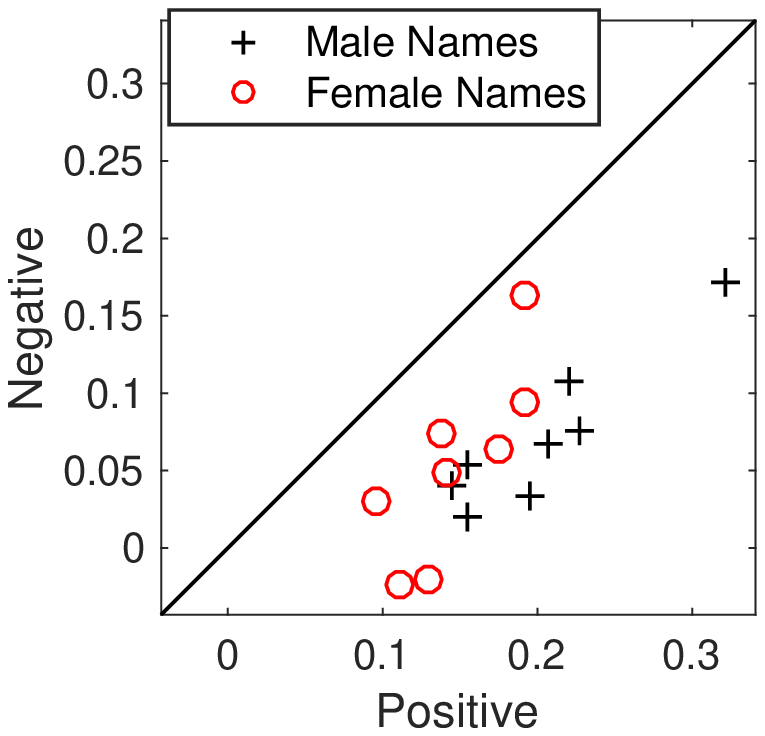}
        \caption{{\centering Association of Gender with Sentiment}\label{fig:MFPosNeg}}
    \end{subfigure}
    ~
    \begin{subfigure}[t]{0.3\textwidth}
        \centering
        \includegraphics[width=\textwidth,height=\textwidth]{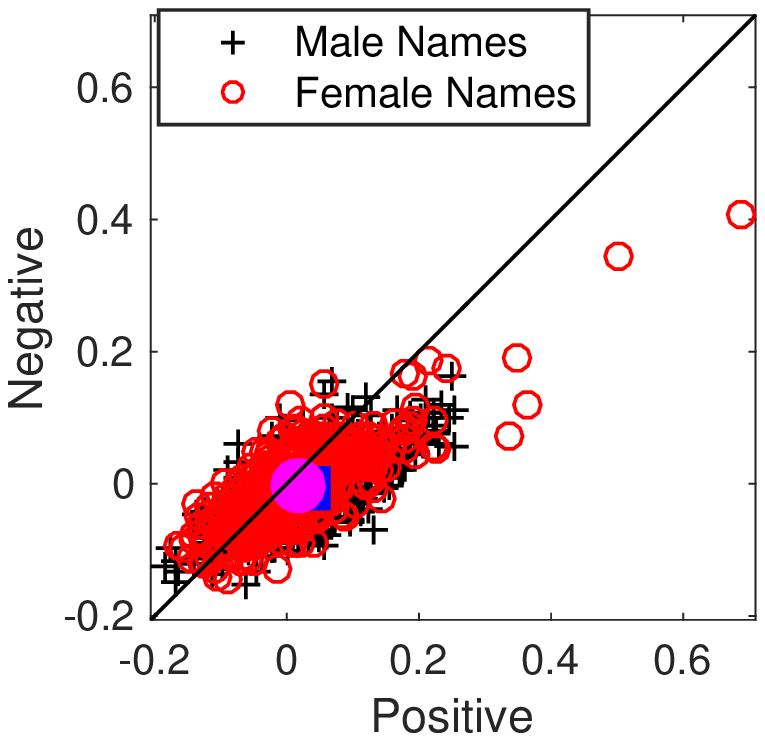}
        \caption{{\centering Extended Association of Gender with Sentiment}\label{fig:BabyPosNeg}}
    \end{subfigure}
    \caption{Association between different words and concepts in Experiment 1, resulting from the proposed LIWC Word Embedding Association Test.}
\end{figure*}

\subsection{Experiment 1: LIWC Word Embedding Association Test (LIWC-WEAT)\label{sec:BrysonTest}}
In this experiment, we introduce the LIWC Word Embedding Association Test (LIWC-WEAT), where we measure the association between sets of target words with larger sets of words known to relate to sentiment and gender coming from the LIWC lexica \cite{pennebaker2001linguistic}. We begin by using the target words from \cite{caliskan2017semantics} which were originally used in \cite{greenwald1998measuring}, allowing us to directly compare our findings with the original WEAT.

%In this experiment, we attempted to replicate the findings of \cite{caliskan2017semantics}, in which they aim to replicate the Implicit Association Test (IAT) \cite{greenwald1998measuring} by taking the same set of target words and measuring their association using GloVe word embeddings. 

Our approach differs from that of \cite{caliskan2017semantics} in that while we use the same set of target words in each test, we use an expanded set of attribute words, allowing us to perform a more rigorous, systematic study of the associations found within the word embeddings. For this, we use attribute words sourced from the LIWC lexica \cite{pennebaker2001linguistic}. The categories specified in the LIWC lexica are based on many factors, including emotions, thinking styles, and social concerns. For each of the original word categories used in \cite{caliskan2017semantics}, we matched them with their closest equivalent within the LIWC categories, for example matching the word lists for `career' and `family' with the `work' and `family' LIWC categories.

We tested the association between each target word and the set of attribute words using the method described in Sec.~\ref{sec:wordComp}, focussing on the differences in association between sentimental terms and European- and African-American names, subject disciplines to each of the genders, career and family terms with gendered names, as well as looking at the association between gender and sentiment.

\subsubsection{Association of European and African-American Names with Sentiment \label{sec:AfriAmer}}
%%%%%%%%%%%%%% FROM HERE (doesn't look very nice also)
Taking the list of target European-American and African-American names used in \cite{caliskan2017semantics}, we tested each of them for their associated with the positive and negative emotion concepts found in \cite{pennebaker2001linguistic} by using the methodology described by Eq.~\ref{eqn:word2set} in Sec.~\ref{sec:wordComp}, replacing the short list of words used to originally represent pleasant and unpleasant attribute sets.

Our test found that while both European-American names and African-American names are more associated with positive emotions than negative emotions, the test showed that European-American names are more associated with positive emotions than their African-American counterparts, as shown in Fig.~\ref{fig:AfriAmer}. This finding supports the association test in \cite{caliskan2017semantics}, where they also found that European-American names were more pleasant than African-American names.

\subsubsection{Association of Subject Disciplines with Gender \label{sec:ArtSciMath}}
A further test was conducted to find the association between words related to different subject disciplines (\textit{e.g.} arts, maths, science) with each of the genders using the `he' and `she' categories from LIWC \cite{pennebaker2001linguistic}.

The results of our test again support the findings of \cite{caliskan2017semantics}, with Maths and Science terms being more closely associated with males, while Arts terms are more closely associated with females, as shown in Fig.~\ref{fig:ArtSciMath}.

\begin{figure*}[t!]
\captionsetup{justification=centering}
    \centering
    \begin{subfigure}[t]{0.3\textwidth}
        \centering
        \includegraphics[width=\textwidth,height=\textwidth]{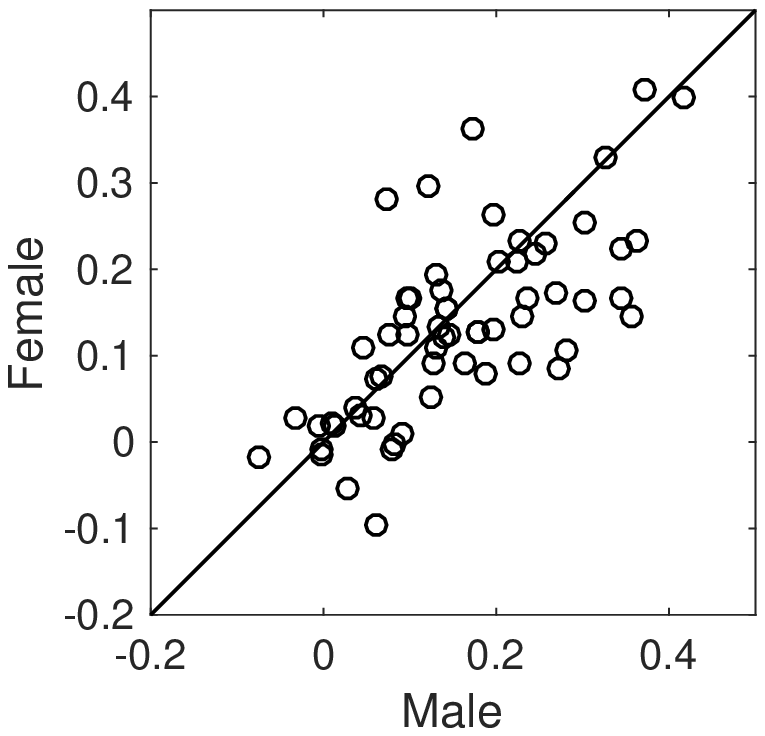}
        \caption{Association of Occupation with Gender\label{fig:jobsMF}}
    \end{subfigure}%
    ~ 
    \begin{subfigure}[t]{0.3\textwidth}
        \centering
        \includegraphics[width=\textwidth,height=\textwidth]{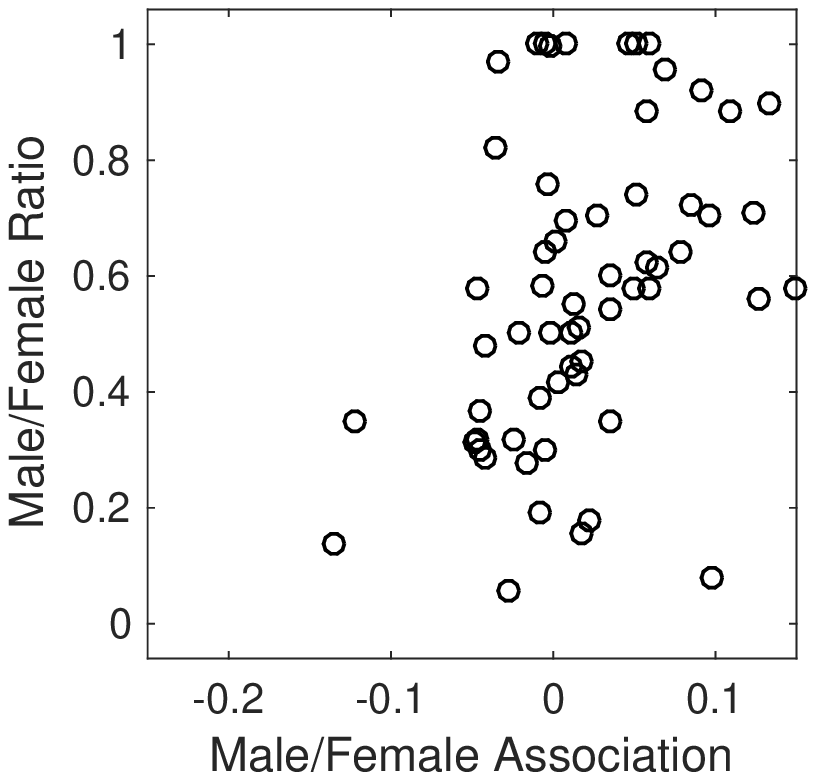}
        \caption{Occupation Statistics versus Gender Association\label{fig:Ratios}}
    \end{subfigure}
    \caption{Results from Experiment 2, showing the association between gender and its relation to the number of men and women working in those roles.}
\end{figure*}

\subsubsection{Association of Gender with Career and Family \label{sec:careerFamily}}
%% For pos / neg emo
Taking the list of target gendered names used in \cite{caliskan2017semantics}, we tested each of them for their associated with the career and family concepts using the categories of `work' and `family' found in LIWC \cite{pennebaker2001linguistic}.

As shown in Fig.~\ref{fig:genderWorkFamily}, we found that the set of male names was more associated with the concept of work, while the female names were more associated with family, mirroring the results found in \cite{caliskan2017semantics}.

Extending this test, we generated a much larger set of male and female target names from an online list of baby names\footnote{Baby names were taken from \url{http://bit.ly/2Dmqjco}, separated into two gendered lists.}. Repeating the same test on this larger set of names, we found that male and female names were much less separated than suggested by previous results, with only minor differences between the two, as shown in Fig.~\ref{fig:babyWorkFamily}.

\subsubsection{Association of Gender with Sentiment \label{sec:PosNeg}}
Extending the number of tests performed in the original WEAT study, we additionally tested the set of target male and female names and computed their association with the positive and negative emotions. We found that both sets of names are considered to be positive, similarly to the European-American and African-American names used in the previous test, but with male names appearing to be slightly more positive, as shown in Fig.~\ref{fig:MFPosNeg}.

We further tested these associations using our extended list of gendered baby names, as in Sec.~\ref{sec:careerFamily}, finding that there is no clear difference between the positive and negative sentiment attached to names of different gender in the word embedding.

\begin{table}[t!]
\begin{center}
\caption{\label{tab:top10}List of the top 10 occupations per gender by their association with gender.}
 \begin{tabular}[\textwidth]{@{}l p{0.85\columnwidth} @{}} 
 \toprule
 Gender & Occupations most associated with a gender\\
 \midrule
 Male & Manager, Engineer, Coach, Executive, Surveyor, Secretary, Architect, Driver, Police, Caretaker, Director\\
 Female & Housekeeper, Nurse, Therapist, Bartender, Psychologist, Designer, Pharmacist, Supervisor, Radiographer, Underwriter\\ 
 \bottomrule
\end{tabular}
\end{center}
\end{table}

\subsection{Experiment 2: Associations between Occupations and Gender}
In this experiment, we test the association between different occupations and gender categories coming from LIWC \cite{pennebaker2001linguistic}. The association between each of the occupations is further contrasted against official employment statistics for the United Kingdom detailing the actual number of people working in each job role.

\subsubsection{Association of Occupation with Gender}
We first generated a list of $62$ occupations from data published by the Office of National Statistics \cite{employStats}, filtering the list to only include those occupations for which there is reliable employment statistics and can be summarised by a single word in the embedding, \textit{e.g.} doctor, engineer, secretary. For each of these occupations, we tested their association with each of the genders, as shown in Fig.~\ref{fig:jobsMF}, with the top ten occupations associated with each gender shown in Table~\ref{tab:top10}. We found there was a $70\%$  ($p\mbox{-value} < 10^{-10}$) correlation in the closeness of association between occupations and each of the gender attribute sets.

\begin{figure*}[t!]
\captionsetup{justification=centering}
    \centering
    \begin{subfigure}[t]{0.3\textwidth}
        \centering
        \includegraphics[width=\textwidth,height=\textwidth]{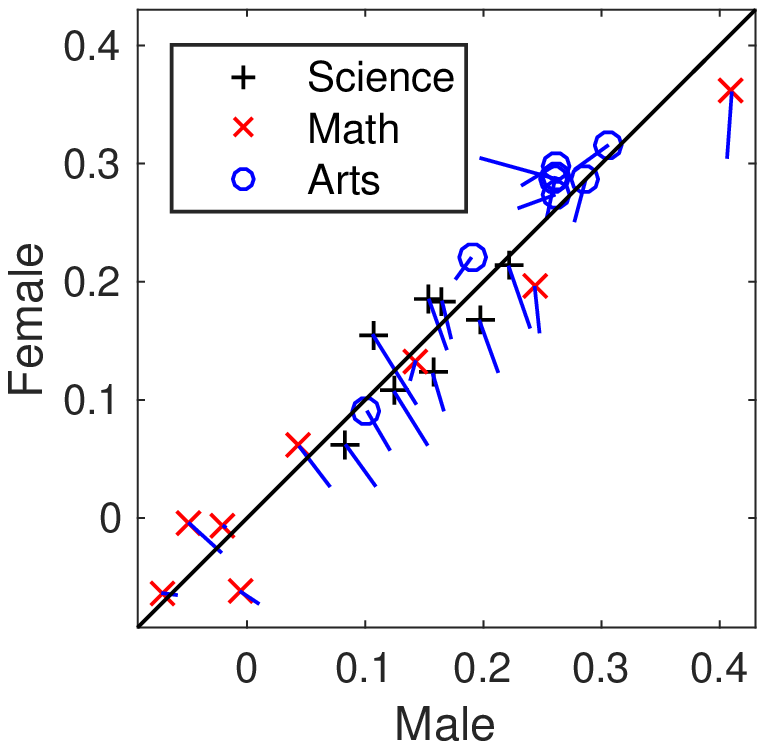}
        \caption{Revised Association of Subject Discipline with Gender\label{fig:ArtSciMathDebias}}
    \end{subfigure}
    ~ 
    \begin{subfigure}[t]{0.3\textwidth}
        \centering
        \includegraphics[width=\textwidth,height=\textwidth]{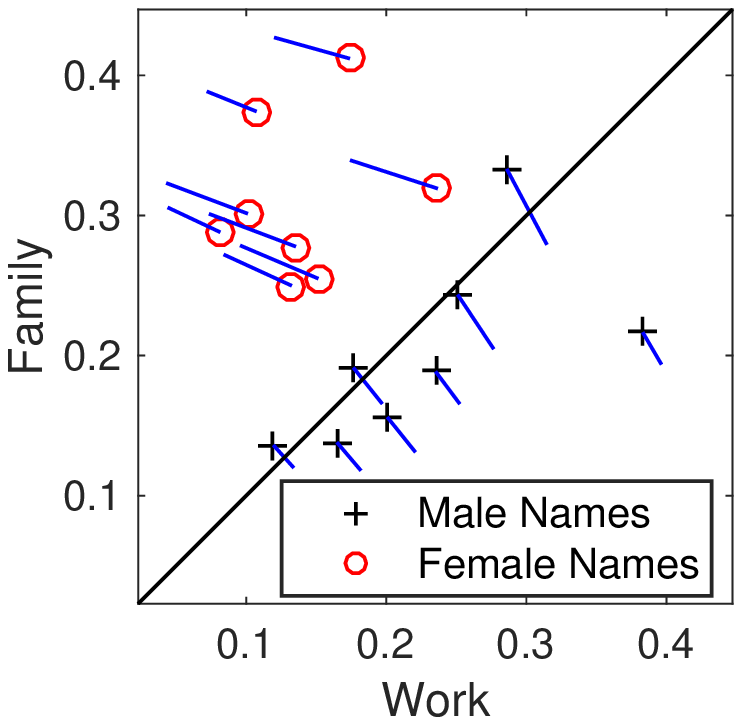}
        \caption{Revised Association of Gender with Career and Family}
    \end{subfigure}
    ~
    \begin{subfigure}[t]{0.3\textwidth}
        \centering
        \includegraphics[width=\textwidth,height=\textwidth]{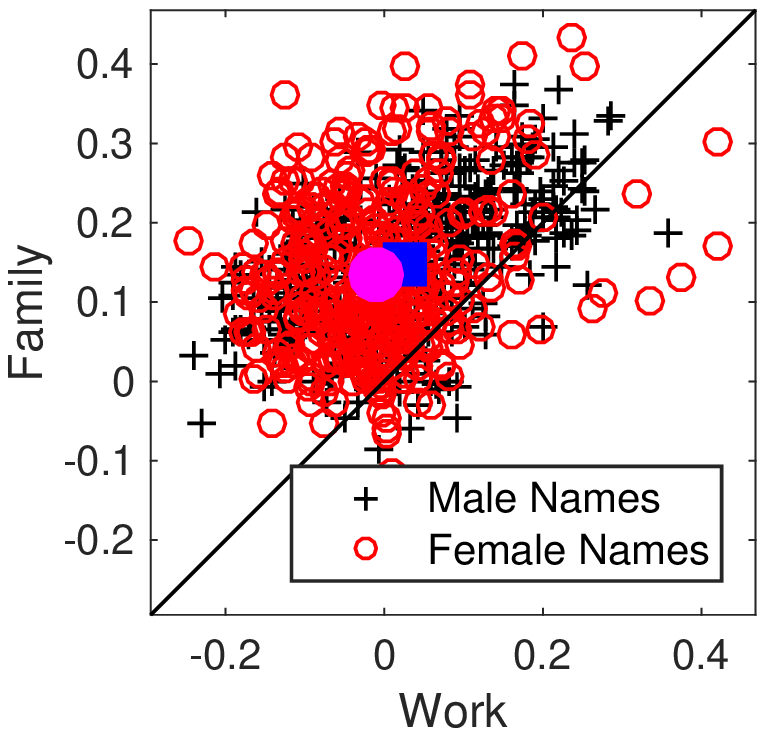}
        \caption{Revised Extended Association of Gender with Career and Family \label{fig:careerFamilyDebias}}
    \end{subfigure}
    \\\vspace{0.65cm}
    \begin{subfigure}[t]{0.3\textwidth}
        \centering
        \includegraphics[width=\textwidth,height=\textwidth]{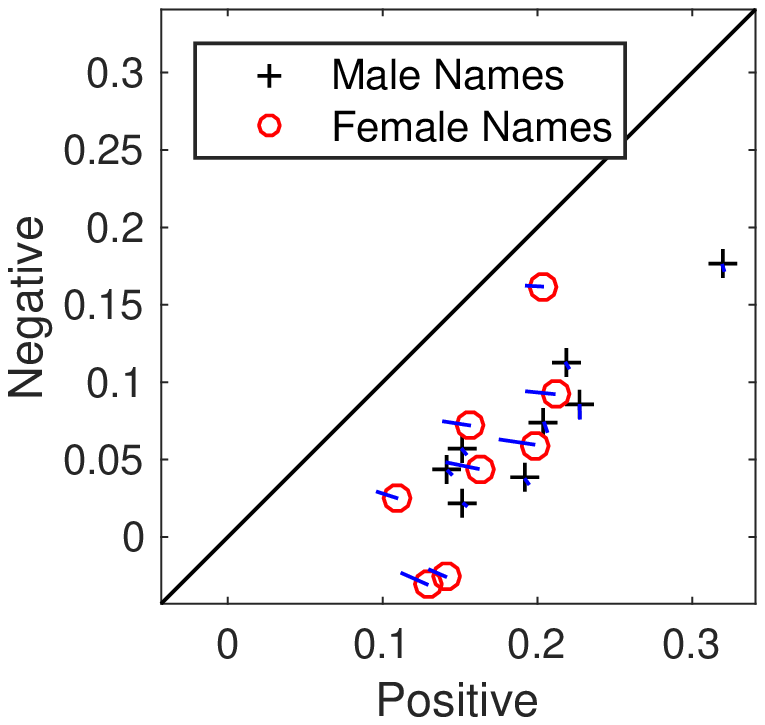}
        \caption{Revised Association of Gender with Sentiment}
    \end{subfigure}
    ~
    \begin{subfigure}[t]{0.3\textwidth}
        \centering
        \includegraphics[width=\textwidth,height=\textwidth]{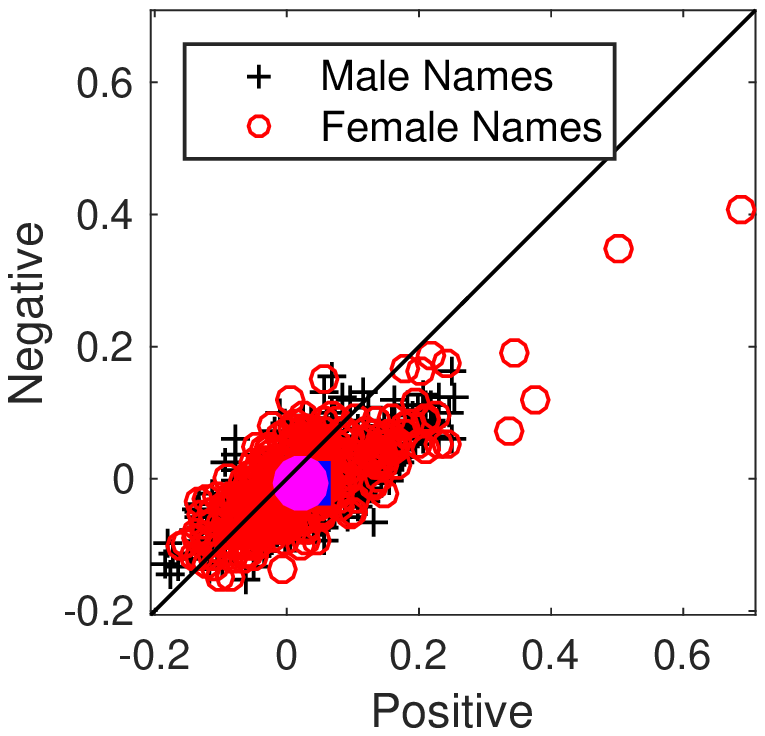}
        \caption{Revised Extended Association of Gender with Sentiment}
    \end{subfigure}
    ~
    \begin{subfigure}[t]{0.3\textwidth}
        \centering
        \includegraphics[width=\textwidth,height=\textwidth]{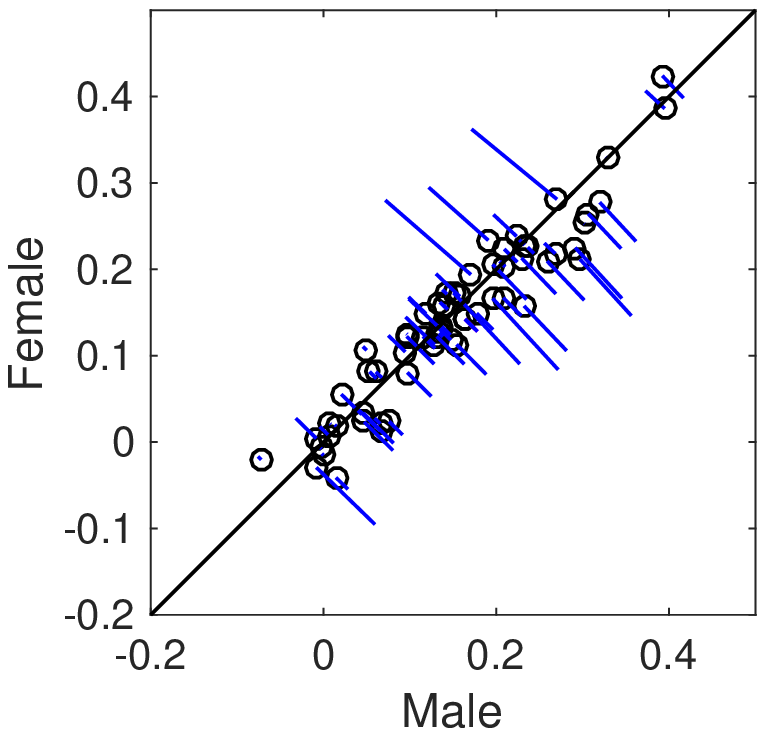}
        \caption{Revised Association of Occupation with Gender\label{fig:jobsDebias}}
    \end{subfigure}%
    \\ \vspace{0.65cm}
    \begin{subfigure}[t]{0.3\textwidth}
        \centering
        \includegraphics[width=\textwidth,height=\textwidth]{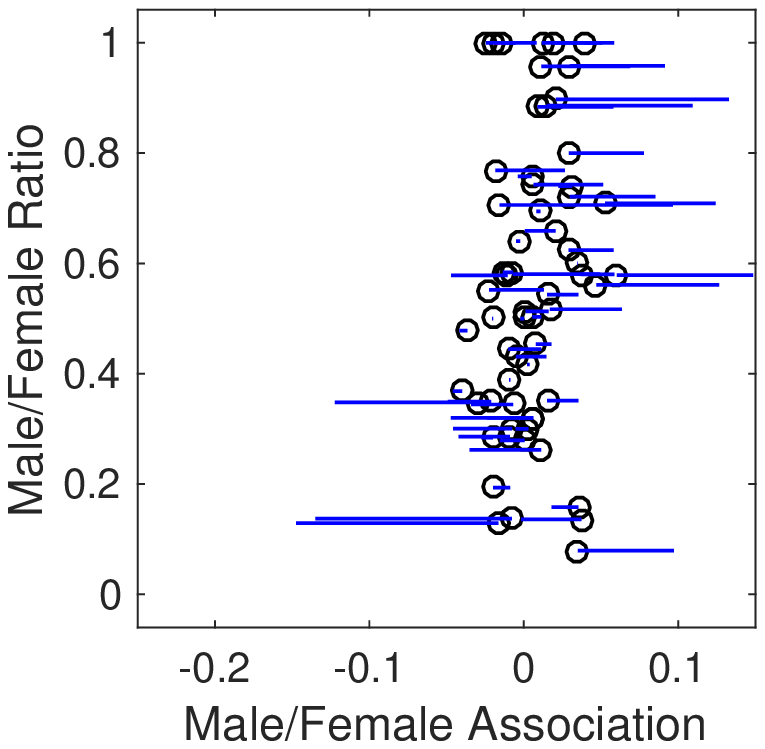}
        \caption{Occupation Statistics versus Revised Gender Association\label{fig:ratioChange}}
    \end{subfigure}
    \caption{Association between different words and concepts in Experiment 3 after word vectors have been debiased via orthogonal projection in the gender direction. Line-traces shown in blue indicate where points have moved from after debiasing.}
\end{figure*}

\subsubsection{Occupation Statistics versus Occupation Association \label{sec:occupation}}
Using the list of occupations from the previous section, we compared their association with each of the genders with the ratio of the actual number of men and women working in those roles, as recorded in the official statistics \cite{employStats}, where $1$ indicates only men work in this role, and $0$ only women. We found that there is a strong, significant correlation ($\rho = 0.57, p\mbox{-value} < 10^{-6}$) between the word embedding association between gender and occupation and the number of people of each gender in the United Kingdom working in those roles. This supports a similar finding for U.S. employment statistics using an independent set of occupations found in \cite{caliskan2017semantics}.

\subsection{Experiment 3: Minimising Associations via Orthogonal Projection}
%Intro to experiment, i.e. what do we do in this experiment, what do we test etc.
In this experiment, we deploy a method for removing bias from word embeddings, first published in \cite{bolukbasi2016man}, and repeat all previous association tests related to gender reported in this paper, empirically showing the effect of bias removal on the word associations.

\subsubsection{Finding an Orthogonal Projection for Gender}
To remove gender from the embedding, we first need to find a projection within the space that best encapsulates the gender differences between words. To find the best projection, we began from a list of $5$ gendered pronouns in LIWC \cite{pennebaker2001linguistic}. For each of the pronouns, we paired them with their gender-opposite, for example pairing ``he'' and ``she'', ``himself'' and ``herself'' and so on. Taking the word vector from the embedding for each pronoun, we computed their difference, as described in Sec.~\ref{sec:removingbias}, giving us a set of 5 potential gender projections.

Each gender projection was tested against an independent set of paired gender words sourced from WordNet \cite{miller1995wordnet}. After applying the gender projection to the test word-pairs, following the procedure of \cite{bolukbasi2016man}, we measured the average difference between the word-pairs. The gender projection that led to the word-pairs that are closest together (smallest difference) was then selected as our gender projection, corresponding to the difference between the vectors for ``himself'' and ``herself''.

\subsubsection{Revised Association Tests}
Using the orthogonal gender projection found in the previous section, we repeated the tests from the LIWC-WEAT in Sec.~\ref{sec:BrysonTest} that were related to gender. This included the association of science, mathematics and the arts with gender, the association of male and females names with sentiment, work and family, and the ranking of occupations by their gender association.

In Experiment 1, we previously found that the disciplines of science and maths were more associated with male terms in the embedding, while the arts were closer to female terms. The association of each of these subject disciplines with gender after orthogonal projection was found to be more balanced, with closer to equal association for both male and female terms, shown in Fig.~\ref{fig:ArtSciMathDebias}.

Male and Females names tested in \cite{caliskan2017semantics} showed a clear distinction in their association with work and family respectively, with our replication of the test in Sec.~\ref{sec:careerFamily} finding the same results. Performing the same tests again after applying the gender projection to both name lists, we wished to quantify the change in associations. We calculated the change in the distance between the centroids of each set of names before and after applying the orthogonal gender projection, finding that the association with work for males and family for females reduced, closing the gap between male and female names by $37.5\%$ for the target names found in the original WEAT and $66\%$ for the extended list of names respectively.

In our experiment looking at the association of positive and negative emotions with male and female names, we found that male and female names were both positive, with male names being slightly more associated with positive emotions than female names. The same finding were also true when using a larger set of names and making the same comparison. Applying the orthogonal gender projection to the word vectors, we again looked at how much the difference between the two sets was reduced. We found that for the target names found in the original WEAT, the distance between the two sets of names was reduced by $27\%$, while for the extended list the difference was reduced by $40\%$.

In Experiment 2, we found that there was a significant correlation of $70\%$  between the male and female association of each occupation, while comparing the associations with official statistics of the number of men and women in each role showed a correlation of $53\%$. Again, applying the orthogonal gender projection and repeating these tests, we found that, on average, occupations moved closer to having an equal association with each of the genders (Fig.~\ref{fig:jobsDebias}) and that their association with gender was not significantly correlated ($\rho = 0.178, p\mbox{-value} = 0.167$) with the number of men and women working in each role.

\section{Discussion}
In our experiments, we have shown the effect of one debiasing procedure for reducing the association a given word has in a word embedding generated from natural language corpora with concepts related to gender. Being able to do so relies on a set of gendered terms from which we can obtain pairings with opposite meaning, allowing us to find an orthogonal projection within the space. This will not always be possible for every type of bias that we may wish to remove (or at least reduce) in an embedding because there will not always be a suitable word vector pair that can be used to represent a given bias.

Other biases which are present may also be impossible to detect with our LIWC-WEAT method, as a pre-defined and validated list of words from LIWC were required to perform the tests. Other potentially undesired biases such as race or age are not currently able to be captured using the LIWC lexica, and thus different, carefully considered sets of words would need to be curated.

Indeed, general solutions to this problem are probably impossible, for philosophical reasons, but we believe that biases can at least be mitigated or compensated for, by removing specific subtypes of bias, given we have ways to measure and detect them in the first place. However, in this process, care should also be taken as we may introduce or compound other existing biases in the embeddings.

% {\color{blue} This will have an impact on downstream systems that use embeddings. While it is not desirable to make decisions based on prejudices, removing them results in a loss of information. This could result in a reduced performance for downstream systems. Looking at the impact of debiasing on Fig.~\ref{fig:Ratios} and Fig.~\ref{fig:ratioChange} shows an impact of debiasing on the word vectors. While the occupations are now closer to gender neutrality regardless of original biases shown, it now less represents the real world when looking at those occupational statistics with the correlation reducing from $\rho =0.57$ to $\rho = 0.178$. For an downstream system that would benefit from the information provided in gendered bias, this shows that there would be a reduction in performance due to the debiasing of these embeddings. However in comparison, there may be systems that could potentially be required to be unbiased by design so this may prove to be a valid method of ensuring minimal biases for gender.} 

% {\color{blue} While removing biases in word embeddings may be viewed as the removal of information, it may also be seen as changing a systems world view to be more idealistic of what may be required from it. This may not have a benefit to all systems that could potentially use embeddings, however it could provide a fairer system depending on the purpose of its use within an downstream system.}
%Is the priority to correct the embeddings or the society that they reflect
%Would there be an application that would benefit from the corrected embeddings
\section{Conclusions}
If we want AI to take a central position in society, we need to be able to detect and remove any source of possible discrimination, to ensure fairness and transparency, and ultimately trust in these learning systems. Principled methods to measure biases will certainly need to play a central role in this, as will an understanding of the origins of biases, and new  developments in methods that can be used to remove biases once detected.

In this paper, we have introduced the LIWC-WEAT, a set of objective tests extending the association tests in \cite{caliskan2017semantics} by using the LIWC lexica to measure bias within word embeddings. We found bias in both the associations of gender and race, as first described in \cite{caliskan2017semantics}, while additionally finding that male names have a slightly higher positive association than female names. 
Biases found in the embedding were also shown to reflect biases in the real world and the media, where we found a correlation between the number of men and women in an occupation and its association with each set of male and female names. 
Finally, using a projection algorithm \cite{bolukbasi2016man}, we were able to reduce the gender bias shown in the embeddings, resulting in a decrease in the difference between associations for all tests based upon gender.

Further work in this direction will include removing bias in $n$-gram embeddings, embeddings that include multiple languages and new procedures for both generating better projections to remove a given bias, using debiased embeddings as an input to an upstream system and testing performance, and learning word embeddings which can be generated without chosen directions by construction.

\section*{Acknowledgements}
AS is supported by EPSRC Centre for Communications. TLW and NC are support by the FP7 Ideas: European Research Council Grant 339365 - ThinkBIG.

\bibliographystyle{plain}
\bibliography{main}

\end{document}